\documentclass[letterpaper,10pt,conference]{ieeeconf}
\IEEEoverridecommandlockouts
\usepackage{amsmath,amssymb}
\usepackage{tensor}
\usepackage{graphicx}
\usepackage{caption}
\usepackage{subcaption}
\usepackage{multirow}
\usepackage{overpic}
\usepackage{rotating}
\usepackage{cite}
\usepackage{verbatim}
\graphicspath{{./fig/}}
\usepackage{algorithm}
\usepackage{algorithmic}
\usepackage{color}
\usepackage{soul}
\usepackage{float}
\usepackage{bm}
\usepackage{xspace}
\usepackage{flushend}
\usepackage{tikz}

\DeclareMathOperator*{\argmax}{arg\!\max}

\def\GG{\mathcal{G}}
\def\KK{\mathcal{K}}\def\LL{\mathcal{L}}
\def\NN{\mathcal{N}}
\def\PP{\mathcal{P}}
\def\SS{\mathcal{S}}

\def\Cbb{\mathbb{C}}
\def\Ebb{\mathbb{E}}

\def\Rbb{\mathbb{R}}

\def\R{\Rbb}\def\C{\Cbb}

\def\GP{\GG\PP}
\def\der{\mathrm{d}}

\newcommand{\dr}{\text{d}}

\def \algo {PIPC\xspace}

\def \mdpol {MDP-OL\xspace}
\def \mdpcl {MDP-CL\xspace}
\def \pomdpol {POMDP-OL\xspace}
\def \pomdpcl {POMDP-CL\xspace}

\newtheorem{lemma}{Lemma}

\title{\LARGE \bf Approximately Optimal Continuous-Time Motion Planning and Control\\ via Probabilistic Inference}

\author{Mustafa Mukadam, Ching-An Cheng, Xinyan Yan, and Byron Boots
\thanks{Mustafa Mukadam, Ching-An Cheng, Xinyan Yan, and Byron Boots are affiliated with the Institute for Robotics and Intelligent Machines, Georgia Institute of Technology, Atlanta, GA 30332, USA. {\tt \{mmukadam3, cacheng, xyan43\}@gatech.edu}, {\tt bboots@cc.gatech.edu}.}
}

\begin{document}
\maketitle
\thispagestyle{empty}
\pagestyle{empty}

%%%%%%%%%%%%%%%%%%%%%%%%%%%%%%%%%%%%%%%%%%%%%%%%%%%%%%%%%%%%%%%%%%%%%%%%%%%%%%%%%%%%%%%%%%%%%%%%%%%%%%%%%%%%%%%%

\begin{abstract}
The problem of optimal motion planing and control is fundamental in robotics. However, this problem is intractable for continuous-time stochastic systems in general and the solution is difficult to approximate if non-instantaneous nonlinear performance indices are present. In this work, we provide an efficient algorithm, \algo (Probabilistic Inference for Planning and Control), that yields approximately optimal policies  with arbitrary higher-order nonlinear performance indices. Using probabilistic inference and a Gaussian process representation of trajectories, \algo exploits the underlying sparsity of the problem such that its complexity scales linearly in the number of nonlinear factors. We demonstrate the capabilities of our algorithm in a receding horizon setting with multiple systems in simulation.
\end{abstract}

%%%%%%%%%%%%%%%%%%%%%%%%%%%%%%%%%%%%%%%%%%%%%%%%%%%%%%%%%%%%%%%%%%%%%%%%%%%%%%%%%%%%%%%%%%%%%%%%%%%%%%%%%%%%%%%%

\section{Introduction}

A fundamental goal in robotics is to efficiently compute trajectories of {actions} that drive a robot to achieve some {desired behavior}. We seek a control policy in a multi-stage decision problem~\cite{bertsekas1995dynamic} that can maximize performance indices that describe, for example, the smoothness of motion, energy consumption, or the likelihood of avoiding an obstacle. 

Hierarchical planning and control has been used to solve this problem in practice~\cite{arkin1998behavior}. The idea is to first generate a desired state sequence~\cite{kavraki1996probabilistic,kuffner2000rrt,byravan2014space,schulman2014motion,Mukadam-ICRA-16,Marinho-RSS-16,Dong-RSS-16} without considering full system dynamics, and then design a robust low-level controller for tracking. Because the dynamic constraints are relaxed, it becomes possible for an algorithm to plan a trajectory that satisfies complicated, higher-order performance indices~\cite{toussaint2014newton,Marinho-RSS-16,Dong-RSS-16}, offering greater flexibility in system design. Sampling-based planning techniques can even provide formal guarantees such as probabilistically complete solutions~\cite{kavraki1996probabilistic,kuffner2000rrt}. However, recent work has started to challenge this classical viewpoint by incorporating more dynamic constraints within trajectory planning in search of solutions with improved optimality~\cite{lavalle2001randomized,tedrake2010lqr}. 

\begin{figure}[!t]
	\centering
	\begin{subfigure}[b]{0.41\textwidth}
		\centering
		\begin{tikzpicture}
		\node[anchor=south west,inner sep=0] (image) at (0,0) {\includegraphics[trim={20 30 220 0},clip,width=0.98\linewidth]{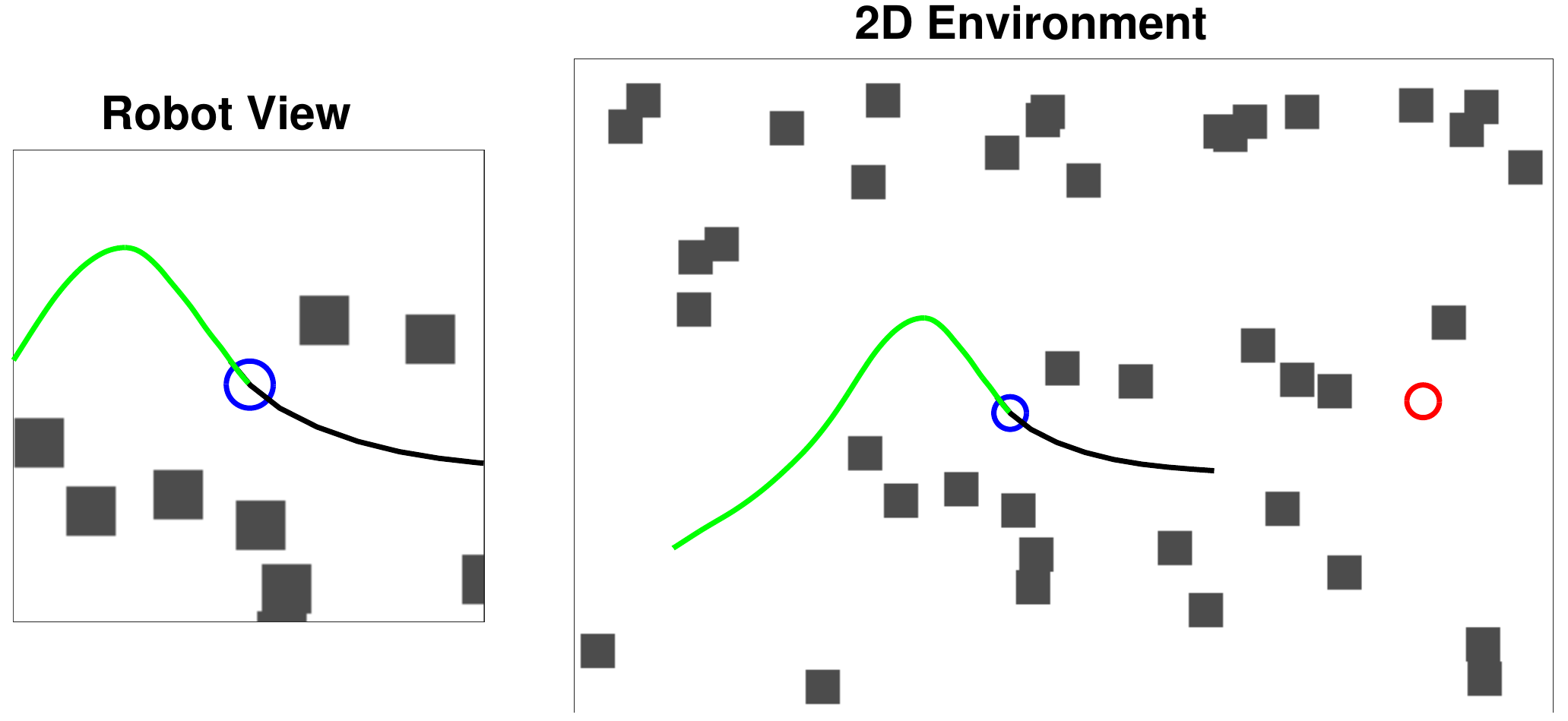}};
		\begin{scope}[x={(image.south east)},y={(image.north west)}]
		\draw[black, thick] (0.01,0.14) rectangle (0.31,0.79);
		\draw[black, thick] (0.365,0.01) rectangle (0.99,0.92);
		\end{scope}
		\end{tikzpicture}
		\caption{}
	\end{subfigure}
	\begin{subfigure}[b]{0.2\textwidth}
		\centering
		\includegraphics[trim={320 80 250 0},clip,width=0.98\linewidth]{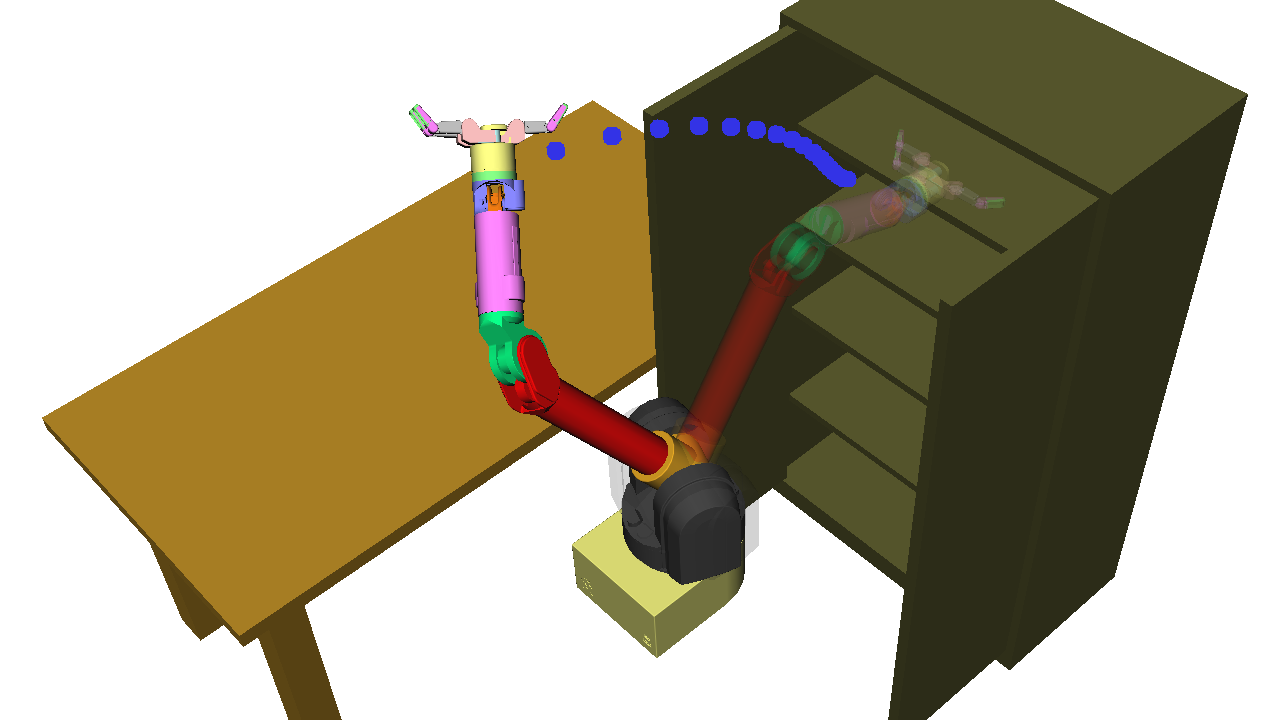}
		\caption{}
	\end{subfigure}
	\begin{subfigure}[b]{0.2\textwidth}
		\centering
		\includegraphics[trim={300 150 450 60},clip,width=0.9\linewidth]{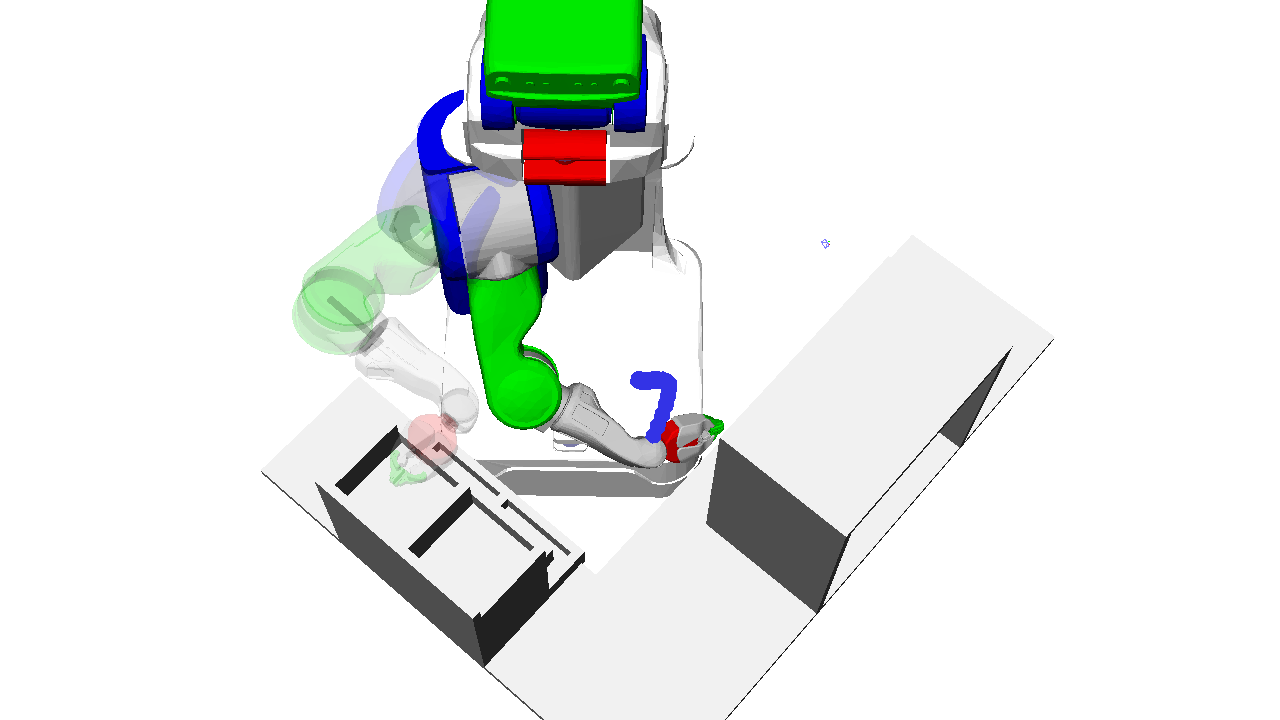}
		\caption{}
	\end{subfigure}
	\caption{\algo used on (a) a 2D holonomic robot (blue) to reach goal (red) in a 2D environment with dynamic obstacles, where executed trajectory is in green and current planned horizon is in black, (b) a 7-DOF WAM arm, and (c) a  PR2's right arm where the semitransparent arm is the goal configuration and dotted blue end effector trajectory is the current planned horizon.}
	\label{fig:misc}
	\vspace{-5mm}
\end{figure}

A  theoretically elegant approach  would be to address both the planning and control problems within a stochastic optimal control framework. Unfortunately, since the states and actions are coupled through system dynamics, exact solutions become intractable with the exception of simple cases known as linearly solvable problems~\cite{kappen2005linear}.\footnote{Affine systems with quadratic instantaneous control cost, or fully controllable discrete-time systems.}

These challenges have motivated researchers to find approximate solutions rather than directly approximating the original problems with hierarchical approaches. 
One simple approach is direct policy search~\cite{levine2013guided,deisenroth2011pilco}, which uses first-order information to find a locally optimal policy. To improve the convergence rate, differential dynamic programming (DDP) has been widely adopted as the foundation of locally optimal algorithms~\cite{mayne1966second,todorov2005generalized,todorov2009iterative}, which solve local linear-quadratic Gaussian (LQG) subproblems and iteratively improve these suboptimal solutions. However, for continuous-time systems, these algorithms would require inefficient high-frequency sampling to construct the LQG subproblems, \emph{even} when the given problem is close to a LQG (e.g. a performance index with only a small set of nonlinear factors, or dynamics with a small amount of nonlinearity). Compared with the hierarchical approach, these algorithms impose a strict structural assumption: they are only applicable to problems that measure performance as an integral of instantaneous functions. 

In this paper, we propose a novel approximately optimal approach to continuous-time motion planning and control that can handle costs expressed as arbitrary higher-order nonlinear factors and exploit a problem's underlying \emph{sparse} structure. Specifically, we consider problems with a performance index expressed as the product of an exponential-quadratic factor for instantaneous costs and a finite number of possibly higher-order nonlinear factors, and provide an algorithm that has linear complexity in the number of nonlinear factors. Moreover, we show the approximately optimal policy can be computed by posterior inference on a probabilistic graphical model, which is a dual to the performance index.

We convert these theoretical results into a practical algorithm called Probabilistic Inference for Planning and Control (\algo) that recursively updates the approximately optimal policy as more information is encountered. To evaluate our approach, we employ \algo on both Markov decision processes (MDPs) and partially-observable MDPs (POMDPs) in dynamic environments with multiple simulated systems (see Fig. \ref{fig:misc}).

%%%%%%%%%%%%%%%%%%%%%%%%%%%%%%%%%%%%%%%%%%%%%%%%%%%%%%%%%%%%%%%%%%%%%%%%%%%%%%%%%%%%%%%%%%%%%%%%%%%%%%%%%%%%%%%%
\subsection{Related Work} \label{sec:Related Work}

Our algorithm contributes to a growing set of research that seeks to reframe planning and control problems as probabilistic inference~\cite{attias2003planning}. Work in this area has formed a new class of approximately optimal algorithms that leverage tools from approximate probabilistic inference, including expectation propagation~\cite{toussaint2009robot} and expectation maximization~\cite{toussaint2006probabilistic,levine2013variational}. A common framework based on KL-minimization~\cite{Kappen2012,rawlik2012stochastic} summarizes the above algorithms as well as approaches like path-integral control~\cite{kappen2005linear}.

We contribute to this field in the following ways. First, we extend the performance index for control algorithms to incorporate nonlinear factors with arbitrary higher-order connections in time. In contrast to our approach, the  methods mentioned above generally assume that the performance indices factor into instantaneous terms, and thus require dense sampling to solve continuous-time problems. Second, we provide a new approach to derive a Gaussian approximation based on Laplace approximation and Gaussian processes. Third, we define a new class of optimal control problems, called gLEQG (generalized Linear-Exponential-Quadratic-Gaussian), that are solvable after being transformed into their dual probabilistic representation. In particular, we show that gLEQG admits a solution given by posterior inference. This theoretical result, discussed in Section~\ref{sec:optimal policy}, closes the gap in the duality between optimal control and inference. 

This rest of the paper is structured as follows. We begin in Section \ref{sec:problem} by defining the objective function in joint planning and control problems. Then, in Section \ref{sec:Approx Inf}, we present our main results in approximately  optimal motion planning and control. In Section \ref{sec:pi-PAC}, these theoretical results are summarized into an online algorithm \algo that can perform simultaneous planning and control for partially observable stochastic linear systems in dynamic environments. To validate our algorithm, we present the implementation details and experimental results in Section \ref{sec:imp} and Section \ref{sec:eval}. Finally, Section \ref{sec:conc} concludes the paper.  

%%%%%%%%%%%%%%%%%%%%%%%%%%%%%%%%%%%%%%%%%%%%%%%%%%%%%%%%%%%%%%%%%%%%%%%%%%%%%%%%%%%%%%%%%%%%%%%%%%%%%%%%%%%%%%%%

\section{The Problem of Motion Planning and Control} \label{sec:problem}
We begin by introducing some notation.
Let $x_t$, $u_t$, and $z_t$ be the state, action, and observation of a continuous-time partially-observable system at time $t$, and let $\bm{h}_t  = \{z_0, u_0, z_{\delta t}, \cdots, z_t  \} $ be the history of observations and actions until time $t$.\footnote{Here we assume the measurements $z_t$ are taken in discrete time at time $t$ with sampling interval $\delta t$, and $u_t$ is a constant continuous-time trajectory  in time $[t, t+\delta t)$.} 
As shorthand, we use boldface to denote the time trajectory of a variable, and  $\bm{\pi}(\bm{u} | \bm{h})$ to denote the collection of time-varying causal (stochastic) policies $\pi_t(u_t | \bm{h}_t )$ for all $t$.  

We formulate the motion planning and control problem as a finite-horizon stochastic optimization problem over $\bm{\pi}$. Let  $p_{\bm{\pi}}$ be the distribution of $\bm{x}$ and $\bm{u}$ under the stochastic policy $\bm{\pi}$ and system dynamics, and $\SS$ be a finite set of time indices. Here the goal is to find an optimal policy $\bm{\pi}$ to maximize the performance index
\begin{align}
	\max_{\bm{\pi}} J(x_0) = \max_{\bm{\pi}} \Ebb_{ p_{\bm{\pi}} }  \left[ \psi(\bm{x}, \bm{u}) \prod_{S\in\SS} \phi_S ( x_S, u_S ) \right].   \label{eq:optimization}
\end{align}
The objective function in~\eqref{eq:optimization} is defined as the expectation of the  product of two types of factors: a Gaussian process factor $\psi(\cdot)$ and a higher-order nonlinear factor $\phi_S(\cdot)$. These two factors, described below, cover many interesting behaviors that are often desired in planning and control problems. 

%------------------------------------------------------------------------------------------------------------------------------------------------%
\subsection{Higher-order Nonlinear Factors $\phi_S(\cdot) $}
We define factors of the form
\begin{align}
	\phi_S(\cdot) = \exp( - \| f_S(\cdot)\|^2 ), \label{eq:nonlinear factor}
\end{align}
to model nonlinear, higher-order couplings frequently used in planing problems, 
where $f_S(\cdot)$ is a differentiable nonlinear function defined on a finite number of time indices $S \in \SS$.
The structure of $\phi_S(\cdot)$ can model many  performance indices in planning: for example, a simple nonlinear cost function at a single time instance, a penalty based on the difference between the initial and the terminal states/actions, a penalty to enforce consistency across landmarks in time, or the cost of a robot-obstacle collision. As each factor $\phi_S(\cdot)$ depends only on a finite number of states or actions, we refer to the corresponding states $x_S$ and actions $u_S$ as \emph{support} states or \emph{support} actions.

%------------------------------------------------------------------------------------------------------------------------------------------------%
\subsection{Gaussian Process Factors $\psi(\cdot)$} \label{sec:GP factor}
The Gaussian process factor $\psi(\cdot)$ is a generalization of the exponential-of-integral cost function in the optimal control literature~\cite{kumar1981optimal}.
To illustrate, here we consider a special case $\psi(\cdot) = \psi(\bm{u})$. A joint factor between $\bm{x}$ and $\bm{u}$ as in~\eqref{eq:optimization} can be defined similarly.

Let $\GP_u (u_t | {m}^u_t, {\KK}^u_{t,t'})$ be a Gaussian process~\cite{rasmussen2006gaussian}, where $\forall t,t'\in \R$, $\Ebb[u_t]= m^u_t$ and $\C[u_t, u_{t'}]  = {\KK}^u_{t,t'}$. Let $\PP^u_{t,t'}$ be the (positive definite) Green's function of $\KK^u_{t,t'}$ satisfying, $\forall t,t'\in \R$,
$
\delta_{t,t'} = \int \KK^u_{t,s} \PP^u_{s, t' } \der s, \label{eq:Green's function}
$
where $\delta$ is the Dirac delta distribution and the integral is over the length of the trajectory.
We define the Gaussian process factor $\psi(\bm{u})$ as 
\small
\begin{align}
	\psi(\bm{u}) = \exp \left(  - \iint (u_s-{m}^u_s)^T \PP^u_{s, s'} (u_{s'}-{m}^u_{s'})  \der s \der s'\right). \label{eq:gp factor}
\end{align}\normalsize
Loosely speaking, we call~\eqref{eq:gp factor} the \emph{probability}
of a trajectory $\bm{u}$ from 
% belonging to 
$\GP_u (u_t | {m}^u_t, {\KK}^u_{t,t'})$. Note that this notation does not necessarily imply that $\bm{u}$ is a sample path of $\GP_u (u_t | {m}^u_t, {\KK}^u_{t,t'})$; rather, we use~\eqref{eq:gp factor} as a metric between $\bm{u}$ and $\bm{m}^u$. Intuitively, the maximization in~\eqref{eq:optimization} encourages $\bm{u}$ to be close to $\bm{m}^u$ in terms of the distance weighted by $\PP^u_{t,t'}$.

Solving a stochastic optimization problem with~\eqref{eq:gp factor} in the objective function is intractable in general, because $\PP^u_{t,t'}$  is only implicitly defined. However, here we show that when $\GP_u$ is the sum of a Gaussian white noise process and a linearly transformed Gauss-Markov process, the problem is not only tractable but can also extend the classical exponential-of-integral cost to model higher-order behaviors. 

This is realized by defining $\GP_u (u_t | {m}^u_t, {\KK}^u_{t,t'})$ through a linear stochastic differential equation (SDE). Let $y_t$ be the hidden state of $u_t$ (e.g. its higher-order derivatives) and $p(y_0) = \NN(y_0| m^y_0, \KK^y_0)$ be its prior. We set $\GP_u (u_t | {m}^u_t, {\KK}^u_{t,t'})$ as the solution to 
\begin{align}  
	\begin{aligned}
		d y_t &= (D y_t + \eta) dt + G  d \omega \\
		u_t &=  H y_t  +  r_t +  \nu_t
	\end{aligned}
	\label{eq:control LTVSDE}
\end{align}
in which $D$, $\eta$, $G$, $H$  are (time-varying)  system matrices,  $r_t$ is control bias, $d\omega$ is a Wiener process, and $\nu_t$ is a Gaussian white noise process $\GP_\nu(0, Q_\nu \delta_{t,t'})$.
In other words, the Gaussian process $\GP_u (u_t | {m}^u_t, {\KK}^u_{t,t'})$ has mean and covariance functions:
\begin{align}
	m^u_t &= r_t + H m^y_t   \label{eq:SDE sol mean} \\ 
	\KK^u_{t,t'} &=  Q_\nu \delta_{t,t'} + H \KK^y_{t,t'} H^T \label{eq:SDE sol cov}
\end{align}
in which $\GP_y(  m^y_t,  \KK^y_{t,t'})$ is another Gaussian process with 
\begin{align} 
	m^y_t  &=  \Phi_y(t,t_0) m^y_0 + \int_{t_0}^t \Phi_y(t,s) \eta_s \dr s  \label{eq:GP mean}  \\
	\KK^y_{t,t'} &=   \Phi_y(t,t_0) \KK^y_0 \Phi_y(t', t_0)^T + \nonumber \\
	& \qquad \int^{\min(t,t')}_{t_0} \Phi_y(t,s) G_s G_s^T \Phi_y(t',s)^{T} \dr s  \label{eq:GP cov}
\end{align}
and $\Phi_y(t,s)$ is the state transition matrix from $s$ to $t$ with respect to $D$. 
For derivations, please refer to~\cite{sarkka2013spatiotemporal} and therein.

The definitions~\eqref{eq:SDE sol mean} and~\eqref{eq:SDE sol cov} contain the exponential-of-integral cost~\cite{kumar1981optimal}
\[\psi(\bm{u}) = \exp \left( -\int (u_s - r_s)^T Q_\nu^{-1} (u_s- r_s)  \der s \right)\] 
as a special case, which can be obtained by setting $H = 0$ (i.e. $\PP_{t,t'}^u = Q_\nu^{-1}$). In general, it assigns the action $\psi(\bm{u}) $ to be close to $\bm{r}$, even in terms of higher-order derivatives (or their hidden states). This leads to a preference toward smooth control signals. By extension, a joint factor between $\bm x$ and $\bm u$ would also encourage smooth state trajectories (i.e. smaller higher-order derivatives of the state).

Constructing the Gaussian process factor by SDE results in one particularly nice property: If we consider the joint Gaussian process of $y_t$ and $u_t$, then its Green's function is \emph{sparse}. To see this, let $\theta_t = (u_t, y_t)$ and $\bm{\theta} = \{ \theta_1, \theta_2, \dots, \theta_N\}$ and define $\psi(\bm{\theta})$ as its Gaussian process factor similar to~\eqref{eq:gp factor}. Then the double integral in $\psi(\bm{\theta})$ can be broken down into the sum of smaller double integrals, or factorized as
\begin{align}
	\psi(\bm{\theta}) = \tilde\psi(\theta_0)  \prod_{i=1}^{N-1} \tilde\psi(\theta_i, \theta_{i+1}) \label{eq:sparse Markov}
\end{align}
where $\tilde\psi(\cdot)$ has a similar exponential-quadratic form  but over a smaller time interval $[t_i, t_i + \delta t]$. In other words, if we treat each $\theta_i$ as a coordinate, then the exponent of $\psi(\bm{\theta})$ can be written as a quadratic function with a tridiagonal Hessian matrix (please see~\cite{sarkka2013spatiotemporal} for details). This sparse property will be the foundation of the approximation procedure and algorithm proposed in Section~\ref{sec:Approx Inf} and~\ref{sec:pi-PAC}.

%%%%%%%%%%%%%%%%%%%%%%%%%%%%%%%%%%%%%%%%%%%%%%%%%%%%%%%%%%%%%%%%%%%%%%%%%%%%%%%%%%%%%%%%%%%%%%%%%%%%%%%%%%%%%%%%

\section{Approximate Optimization as Inference} \label{sec:Approx Inf}

The mixed features from both planning and control domains in
\eqref{eq:optimization} present two major challenges: the optimization over continuous-time trajectories and the higher-order, nonlinear factors $\phi_S(\cdot)$. The former results in an infinite-dimensional problem, which often requires a dense discretization. The latter precludes direct use of algorithms based on Bellman's equation, because the factors may not factorize into instantaneous terms. 

In this work, we propose a new approach inspired by approximate probabilistic inference. The goal here is to derive an approximation to the problem in \eqref{eq:optimization}, in the form
\begin{align}
	\max_{\bm{\pi}} \Ebb_{ \hat{p}_{\bm{\pi}} } \left [ \psi(\bm{x}, \bm{u}) \prod_{S\in\SS} \hat{\phi}_S ( x_S, u_S ) \right  ],  \label{eq:LEQG approx}
\end{align}
where $\hat{\phi}_S ( \cdot)$ is a local exponential-quadratic approximation of $\phi_S (\cdot)$ and $\hat{p}_{\bm{\pi}}$ is a Gaussian process approximation of $p_{\bm{\pi}}$. We call the problem in~\eqref{eq:LEQG approx} ``gLEQG'' as it generalizes LEQG (Linear-Exponential-Quadratic-Gaussian)~\cite{kumar1981optimal} to incorporate higher-order exponentials in the form of~\eqref{eq:gp factor}. 

In the rest of this section, we show how gLEQG can be derived by using the probabilistic interpretation~\cite{toussaint2009robot} of the factors in~\eqref{eq:optimization}. Further, we show this problem can be solved in linear time $O(|\SS|)$ and its solution can be written in closed-form as posterior inference.

\subsection{Probabilistic Interpretation of Factors} \label{sec:probilistic interpretation}

We begin by representing each factor in~\eqref{eq:optimization} with a probability distribution~\cite{toussaint2006probabilistic}. First, for $\phi_S(\cdot)$, we introduce additional fictitious observations $e_S$ such that $p(e_S| x_S, u_S) \propto \phi_S ( x_S, u_S )$. These new variables $e_S$ can be interpreted as the events that we wish the robot to achieve and whose likelihood of success is reflected proportionally to $\phi_S(\cdot)$. 
Practically, they help us keep track of the message propagation over the support state/action in later derivations.
Second, we rewrite the Gaussian process factor $\psi(\bm{u})$ to include the hidden state $y_t$ in~\eqref{eq:control LTVSDE}, as a joint Gaussian process factor $q(\bm{u},\bm{y})$.\footnote{This step can be carried similarly as the construction of $\psi(\bm{u})$.} With the introduction of $y_t$, the joint Gaussian process $q(\bm{u},\bm{y})$ has the sparse property in~\eqref{eq:sparse Markov} that we desired.

Now, we rewrite the stochastic optimization~\eqref{eq:optimization} in the new notation. Let $\bm{e}_\SS = \{ e_S\}_{S\in\SS}$ and $\xi = (x,y,u)$, and let $p(\bm{x}|\bm{u})$ and $p(\bm{z}|\bm{x})$ be the conditional distributions defined by the system dynamics and the observation model, respectively. 
It can be shown that~\eqref{eq:optimization} is equivalent to 
\begin{align}
	\max_{ \bm{\pi} }  \int  q(\bm{z}, \bm{\xi} | \bm{e}_\SS)  \bm{\pi}(\bm{u} | \bm{h}) \der \bm{\xi}   \der \bm{z} \label{eq:optimization var}
\end{align}
in which we define a joint distribution
\begin{align}
	q(\bm{z}, \bm{\xi},  \bm{e}_\SS)  =   q(\bm{\xi})  p(\bm{z}|\bm{x}) \prod_{S\in\SS} p(e_S | x_S, u_S )  \label{eq:gaussian latent model}
\end{align}
with likelihoods $p(\bm{z}|\bm{x})$ and $p(e_S | x_S, u_S )$, and a  prior on the continuous-time trajectory $\bm{\xi}$
\begin{align}
	q(\bm{\xi}) = p(\bm{x} | \bm{u} )  q(\bm{u}, \bm{y}). \label{eq:joint dist}
\end{align}
Before proceeding, we clarify the notation we use to simplify writing. We use $q$ to denote the \emph{ad hoc} constructed Gaussian process factor (e.g. in~\eqref{eq:gp factor}) and use $p$ to denote the probability distribution associated with the real system. As such, $q$ does not always define an expectation, so the integral notation (e.g. in~\eqref{eq:optimization var}) denotes the expectation over $p$ and $\pi$ that are well-defined probability distributions. But, with some abuse of notation, we will call them both Gaussian processes, since our results depend rather on their algebraic form.

\subsection{Gaussian Approximation} \label{sec:Deriving Gaussian Approximation}

Let $\bm{\xi}_\SS =\{ \xi_S\}_{ S\in \SS  }$ and $\bar{\bm{\xi}}_{\SS} = \bm{\xi} \backslash \bm{\xi}_\SS$.
To derive the gLEQG approximation to~\eqref{eq:optimization}, we notice, by~\eqref{eq:gaussian latent model}, $q(\bm{z}, \bm{\xi}  | \bm{e}_\SS)$ in~\eqref{eq:optimization var} can be factorized into
\begin{align}
	q(\bm{z}, \bm{\xi} | \bm{e}_\SS)  =  q(\bm{z},\bar{\bm{\xi}}_{\SS}| \bm{\xi}_{\SS}) q( \bm{\xi}_{\SS} | \bm{e}_\SS )  \label{eq:Markov}
\end{align}
where  have used the Markovian property in Section~\ref{sec:probilistic interpretation} i.e. given $\bm{\xi}_\SS$, $\bm{e}_\SS$ is conditionally independent of other random variables.
Therefore, if $q(\bm{z},\bar{\bm{\xi}}_{\SS}| \bm{\xi}_{\SS})$ and $q( \bm{\xi}_{\SS} | \bm{e}_\SS )$ can be reasonably approximated as Gaussians, then we can approximate~\eqref{eq:optimization} with~\eqref{eq:LEQG approx}.

However, $q(\bm{z},\bar{\bm{\xi}}_{\SS}| \bm{\xi}_{\SS})$  and $q( \bm{\xi}_{\SS} | \bm{e}_\SS )$ have notably different topologies. $q(\bm{z},\bar{\bm{\xi}}_{\SS}| \bm{\xi}_{\SS})$ is a distribution over continuous-time trajectories, whereas $q( \bm{\xi}_{\SS} | \bm{e}_\SS )$ is a density function on finite number of random variables. Therefore, to approximate~\eqref{eq:optimization}, we need to find a Gaussian \emph{process}
$\hat{q}(\bm{z},\bar{\bm{\xi}}_{\SS}| \bm{\xi}_{\SS})$ and a Gaussian \emph{density} $\hat{q}( \bm{\xi}_{\SS} | \bm{e}_\SS )$.

\subsubsection{Gaussian Process Approximation}
\label{sec:Gaussian Process Approximate}
We derive the Gaussian process approximation $\hat{q}(\bm{z},{\bm{\xi}})$ to $q(\bm{z},{\bm{\xi}})$. With this result, the desired conditional Gaussian process $\hat{q}(\bm{z},\bar{\bm{\xi}}_{\SS}| \bm{\xi}_{\SS})$ is given closed-form.

First we need to define the system dynamics $p(\bm{x}|\bm{u})$ and the observation model $p(\bm{z}|\bm{x})$. For now, let us assume that the system is governed by a linear SDE
\begin{align} 
	\begin{aligned} \label{eq:state SDE} 
		& dx = (A x + B u + b)dt  + F dw   \\
		& z = C x +  v
	\end{aligned}
\end{align}
in which $A$, $B$, $b$, $F$, $C$ are (time-varying) system matrices, $dw$ is a Wiener process, and $v$ is Gaussian noise with covariance $Q_v$.
When a prior is placed on $x_0$ (similar to Section~\ref{sec:GP factor}) it can be shown that the solution to~\eqref{eq:state SDE} $p(\bm{x},\bm{z} | \bm{u})=  p(\bm{z} | \bm{x} ) p(\bm{x} | \bm{u} )$ is a Gaussian process. Since $ q(\bm{u}, \bm{y})$ is also Gaussian process, we have a Gaussian process prior on $\bm{z}$ and $\bm{\xi}$:
\begin{align}
	q(\bm{z}, \bm{\xi}) = p(\bm{x},\bm{z} | \bm{u}) q(\bm{u}, \bm{y}), \label{eq:Gaussian prior}
\end{align}
In this case, no approximation is made and therefore $\hat{q}(\bm{z},\bar{\bm{\xi}}_{\SS}| \bm{\xi}_{\SS})  = q(\bm{z},\bar{\bm{\xi}}_{\SS}| \bm{\xi}_{\SS}) $.

In the case of nonlinear systems, one approach is to treat~\eqref{eq:state SDE} as its local linear approximation and derive  $\hat{q}(\bm{z}, \bm{\xi}) = \hat{p}(\bm{x},\bm{z} | \bm{u}) q(\bm{u}, \bm{y}) $, where $\hat{p}(\bm{x},\bm{z} | \bm{u})$ is the solution to the linearized system. Alternatively, we can learn the conditional distribution  $\hat{p}(\bm{x},\bm{z} | \bm{u})$ from data directly through Gaussian process regression~\cite{rasmussen2006gaussian}. 
However, since our main purpose here is to show the solution when $\hat{p}(\bm{x},\bm{z} | \bm{u})$ is available, from now on we will assume the system is linear and given by~\eqref{eq:state SDE}.

\subsubsection{Gaussian Density Approximation}
\label{sec:Gaussian Density Approximate} 

Unlike $q(\bm{z},\bar{\bm{\xi}}_\SS | \bm{\xi}_\SS )$,  the approximation to $q (\bm{\xi}_\SS | \bm{e}_\SS)$ is more straightforward. First, because $q (\bm{\xi}_\SS | \bm{e}_\SS)$ may not be available in closed form,
we approximate $q (\bm{\xi}_\SS | \bm{e}_\SS)$  with $ \tilde{q}(\bm{\xi}_\SS | \bm{e}_\SS) $
\begin{align}
	q (\bm{\xi}_\SS | \bm{e}_\SS) &\propto q (\bm{\xi}_\SS )  \prod_{S\in\SS} p(e_S | x_S, u_S ) \nonumber  \\
	&\approx \hat{q}(\bm{\xi}_\SS)  \prod_{S\in\SS} p(e_S | x_S, u_S ) \propto \tilde{q}(\bm{\xi}_\SS | \bm{e}_\SS) \label{eq:approximate conditional marginal}
\end{align}
where  $\hat{q}(\bm{\xi}_\SS)$ is the marginal distribution of $\hat{q}(\bm{z}, \bm{\xi})$, found in the previous section. 
Given~\eqref{eq:approximate conditional marginal}, we then find a Gaussian approximation $\hat{q}(\bm{\xi}_\SS | \bm{e}_\SS)$  of $\tilde{q}(\bm{\xi}_\SS| \bm{e}_\SS) $ via a Laplace approximation~\cite{bishop2006pattern}.

For the nonlinear factor from~\eqref{eq:nonlinear factor}, a Laplace approximation of $\tilde{q}(\bm{\xi}_\SS| \bm{e}_\SS) $ amounts to solving a nonlinear least-squares optimization problem. Using the sparsity of the structured Gaussian processes defined by SDEs, the optimization can be completed using efficient data structures in $O(|\SS|)$~\cite{Dong-RSS-16}. For space constraints, we omit the details here; please see Appendix A and~\cite{Dong-RSS-16} for details.

\subsubsection{Summary}
The above approximations allow us to approximate~\eqref{eq:gaussian latent model} with a Gaussian distribution
\begin{align}
	\hat{q}(\bm{z}, \bm{\xi},  \bm{e}_\SS)  =   \hat{p}(\bm{z}, \bm{x}|\bm{u})  q(\bm{u},\bm{y}) \prod_{S\in\SS} \hat{p}(e_S | x_S, u_S ).  \label{eq:approximate gaussian latent model}
\end{align}
In~\eqref{eq:approximate gaussian latent model}, $\hat{p}(\bm{z}, \bm{x}|\bm{u})$ is the Gaussian process approximation of the system, which is exact when the system is linear, and $\hat{p}(e_S | x_S, u_S ) $ is proportional to the exponential-quadratic factor $\hat{\phi}_S ( x_S, u_S )$ in~\eqref{eq:LEQG approx}. Moreover, it can be shown that $\hat{q}(\bm{z}, \bm{\xi},  \bm{e}_\SS) $  is a Laplace approximation of  $\hat{p}(\bm{z}, \bm{x}|\bm{u})  q(\bm{u},\bm{y}) \prod_{S\in\SS} {p}(e_S | x_S, u_S )$ in terms of continuous-time trajectory $\bm{z}$ and $\bm{\xi}$.

%------------------------------------------------------------------------------------------------------------------------------------------------%
\subsection{Finding an Approximately Optimal Policy} \label{sec:optimal policy}

Substituting the results in Section~\ref{sec:Deriving Gaussian Approximation} into~\eqref{eq:optimization var}, we have the approximated optimization problem
\begin{align}
	\max_{\bm{\pi} } \int  \hat{q}(\bm{z}, \bm{\xi} | \bm{e}_\SS)  \bm{\pi}(\bm{u} | \bm{h}) \der \bm{\xi} \der \bm{z}.
	\label{eq:LEQG approx II}
\end{align}
By~\eqref{eq:approximate gaussian latent model}, one can show that~\eqref{eq:LEQG approx II} is equivalent to the problem in~\eqref{eq:LEQG approx}, but expressed in probabilistic notation. 

However, by writing the problem probabilistically, we can avoid the algebraic complications arising from attempting to solve the Bellman's equation of~\eqref{eq:LEQG approx}, which, because of higher-order factors, requires additional state expansion. This simplicity is reflected in the optimality condition for~\eqref{eq:LEQG approx II}:
\begin{align}
	\pi_t^*( u_t | \bm{h}_t ) &= \delta(u_t - u_t^*(\bm{h}_t)) \nonumber  \\
	u_t^*(\bm{h}_t) &= \argmax_{u_t}  \int  \hat{q}(\bm{z},\bm{\xi} | \bm{e}_\SS )  \bm{\pi}^*( \bar{\bm{u}}_t | \bm{h}) \der \bm{x} \der \bm{y}  \der \bm{z} \der \bar{\bm{u}}_t  \nonumber \\
	&=\argmax_{u_t} \hat{q}(u_t| \bm{h}_t, \bm{e}_\SS) \label{eq:determinisitc policy}
\end{align}
in which $\bar{\bm{u}}_t $ denotes $\bm{u}\setminus\{u_t\}$ and $\delta$ is Dirac delta distribution. 
From the last equality in~\eqref{eq:determinisitc policy}, we see that the solution to the maximization problem coincides with the mode of the posterior distribution $\hat{q}(u_t| \bm{h}_t, \bm{e}_\SS)$. As a result, the optimal policies for time $t$ can be derived \emph{forward} in time, by performing inference without solving for the future policies first. Please see Appendix B for the proof. 

We call this property the duality between gLEQG and inference.
This result seems surprising, but 
similar ideas can be traced back to the duality between the optimal control and estimation~\cite{todorov2005generalized,toussaint2009robot}, in which the optimal value function of a linear quadratic problem is computed by backward message propagation without performing maximization. 

Compared with previous work, a stronger duality holds here: gLEQG is dual to the inference problem on the \emph{same} probabilistic graphical model defined by the random variables in Section~\ref{sec:probilistic interpretation}. 
This nice property is the result of the use of an exponential performance index, and enables us to handle higher-order factors naturally without referring to~\textit{ad hoc} derivations based on dynamic programming on extended states. 

Our posterior representation of the policy can also be found in~\cite{toussaint2009robot,boularias2008predictive}, or can be interpreted as one step of posterior iteration~\cite{rawlik2012stochastic}.
In~\cite{toussaint2009robot}, this results from the approximation of the optimal value function, but its relationship to the overall stochastic optimization is unclear. In~\cite{boularias2008predictive}, the posterior representation is reasoned from the notion of a predictive policy representation without further justification of its effects on the whole decision process. Here we derive the policy based on the assumption that 
the associated distribution of~\eqref{eq:optimization} can be approximated by a Gaussian \eqref{eq:approximate gaussian latent model}. Therefore, the condition on which the approximate policy remains valid can be more easily understood or even enforced, as discussed later in Section~\ref{sec:algo}.

%%%%%%%%%%%%%%%%%%%%%%%%%%%%%%%%%%%%%%%%%%%%%%%%%%%%%%%%%%%%%%%%%%%%%%%%%%%%%%%%%%%%%%%%%%%%%%%%%%%%%%%%%%%%%%%%

\begin{algorithm*}[!t]
	\caption{Receding Horizon \algo}\label{alg:algo}
	\begin{algorithmic} [1]
		\renewcommand{\algorithmicrequire}{\textbf{Input:}}
		\renewcommand{\algorithmicensure}{\textbf{Output:}}
		\REQUIRE horizon $t_h$, start time $t_0$, initial belief $q(\xi_{t_0})$
		\ENSURE  success/failure
		\WHILE{not STOP\_CRITERIA}
		\STATE $\hat q(\bm\xi_S|\bm e_S, \bm h_{t_i-\delta t}, u_{t_i-\delta t})$ = getLaplaceApprox($t_i$, $t_h$, $q(\xi_t|\bm h_{t_i-\delta t}, u_{t_i-\delta t})$, ENVIRONMENT)
		\FOR{$t \in [t_i, t_{i+1}]$}
		\STATE $z_t$ = makeObservation()
		\STATE $\hat q( \xi_t | \bm h_{t}, \bm e_S)$ = 
		filterPolicy($z_t$, $\hat{q} (\xi_{t-\delta t} | \bm{h}_{t-\delta t}, \bm e_S)$, $\hat q(\bm\xi_S|\bm e_S, \bm h_{t_i-\delta t}, u_{t_i-\delta t})$)
		\STATE executePolicy($u_t = u_t^*(\bm h_t)$)		
		\STATE $q(\xi_{t+\delta t}|\bm h_{t}, u_{t} )$ = filterState($z_t$, $u_t$, $q(\xi_t|\bm h_{t-\delta t}, u_{t-\delta t} )$)
		\ENDFOR
		\ENDWHILE
		\RETURN checkSuccess()
	\end{algorithmic}
\end{algorithm*}

\section{Probabilistic Motion Planning and Control} \label{sec:pi-PAC}

In Section~\ref{sec:Approx Inf}, we show that if $q( \bm{z}, \bm{\xi} |  \bm{e})$ can be approximated well by a Gaussian distribution, the stochastic optimization in~\eqref{eq:optimization} can be approximately solved as posterior inference~\eqref{eq:determinisitc policy}. This representation suggests that the approximately optimal policy can be updated recursively through Kalman filtering.

\subsection{Recurrent Policy Inference as Kalman Filtering} \label{sec:reccurent policy}

The approximately optimal policy in~\eqref{eq:determinisitc policy} can be viewed as the belief about the current action $u_t$ given the history $\bm{h}_t$ and the fictitious events $\bm{e}_\SS$. Here we exploit the Markovian structure underlying $\hat{q}(\bm{z}, \bm{\xi} |  \bm{e}_\SS)$ to derive a recursive algorithm for updating the belief $ \hat{q}(\xi_t | \bm{h}_t, \bm{e}_\SS)$. Given the belief, the policy can be derived by marginalization.
First,  for $t=0$, we write
\begin{align*}
	\hat{q}(  \xi_0 | \bm{h}_0 , \bm{e}_\SS  ) &\propto p( z_0 | \xi_0 ) \hat{q}(  \xi_0 | \bm{e}_\SS  )  \\
	&= q( z_0 | \xi_0 ) \int q( \xi_0 | \bm{\xi}_{\SS}) \hat{q}( \bm{\xi}_{\SS} | \bm{e}_\SS  ) \der \bm{\xi}_\SS
\end{align*}
in which $q( z_{t} | \xi_{t} ) = p( z_{t} | x_{t} )$ and $q( \xi_0 | \bm{\xi}_{\SS})$ is the conditional distribution defined by~\eqref{eq:Gaussian prior}. After initialization, the posterior $\hat{q}( \xi_t | \bm{h}_t, \bm{e}_\SS )$ can be propagated through prediction and correction, summarized together in one step as 
\begin{align}
	& \hat{q}( \xi_{t+\delta t} | \bm{h}_{t+\delta t}, \bm{e}_\SS )\propto p( z_{t+\delta t} | \xi_{t+\delta t} ) \hat{q}(  \xi_{t+\delta t} | \bm{e}_\SS  ) \nonumber \\
	&= q( z_{t+\delta t} | \xi_{t+\delta t} ) \int \hat{q}( \xi_{t+\delta t} | \xi_{t}, \bm{e}_\SS) \hat{q}( \xi_t | \bm{h}_t, \bm{e}_\SS ) \der \xi_t \label{eq:GP Kalman filtering}
\end{align}
in which the transition is given by
\begin{align}
	\hat{q}( \xi_{t+\delta t} | \xi_{t}, \bm{e}_\SS) &\propto \hat{q}( \xi_k , \xi_{t+\delta t} |\bm{e}_\SS   ) \nonumber \\
	&= \int   q( \xi_k, \xi_{t+\delta t} |  \bm{\xi}_\SS)   \hat{q}( \bm{\xi}_\SS | \bm{e}_\SS )  \der \bm{\xi}_\SS   \label{eq:transition prob}
\end{align}
and $q( \xi_k, \xi_{t+\delta t} |  \bm{\xi}_\SS) $ is given by~\eqref{eq:Gaussian prior}~\cite{barfoot2014batch,Yan-ISRR-15}. Because of the Markovian structure in~\eqref{eq:sparse Markov}, the integral~\eqref{eq:transition prob} only depends on two adjacent support states/actions of $t$ and can be computed in constant time. 
Note, in $\hat{q}( \xi_t| \bm{h}_{t}, \bm{e}_\SS  )$ in~\eqref{eq:GP Kalman filtering}, the action $u_t$ is actually conditioned on the action taken $u_t^*(\bm{h}_t)$. This notation is adopted to simplify the writing. 

Thus, we can view~\eqref{eq:GP Kalman filtering} as Kalman filtering with transition dynamics $ \hat{q}( \xi_{t+\delta t} | \xi_{t}, \bm{e}_\SS)$ and observation process $q( z_{t} | \xi_{t} )$. This formulation gives us the flexibility to switch between open-loop and closed-loop policies. That is, before a new observation $z_{t+\delta t}$ is available, \eqref{eq:transition prob} provides a continuous-time open-loop action trajectory during the interval $(t,t+\delta t)$. 

\begin{figure}[!t]
	\begin{centering}
		{\includegraphics[width=0.7\columnwidth]{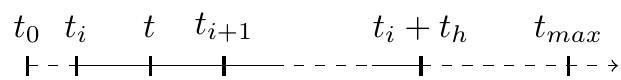}}
		\par\end{centering}
	\protect\caption{Time-line with \algo, where a system that started at $t_0$, is currently at time $t \in [t_i, t_{i+1}]$ between support points $t_i$ and $t_{i+1}$ in $\delta t$ resolution. In a receding horizon setting, $t_+$ represents the receding horizon window $[t_i, t_i + t_h]$, and $t_{max}$ is the (infinite) final time when the algorithm terminates. In a finite horizon setting, $t_h = t_{max} - t_i$.
		\label{fig:time_stamps}}
\end{figure}

This recurrent policy inference is based on the assumption that $\hat{q}( \bm{z}, \bm{\xi} |  \bm{e}_\SS)$ is accurate. 
Although this assumption is not necessarily true in general, it is a practical approximation if the belief about current state $p(x_t | \bm{h}_t)$ is concentrated and the horizon within which~\eqref{eq:determinisitc policy} applies is short.

\subsection{Online Motion Planning and Control} \label{sec:algo}
We now summarize everything into the \algo algorithm. Let $t_i \in \SS$. We compensate for the local nature of~\eqref{eq:determinisitc policy} by recomputing a new Laplace approximation $q( \bm{z}_{t_+}, \bm{\xi}_{t_+} | \bm{h}_t, \bm{e}_\SS)$ whenever $t \in\SS$, and applying filtering to update the policy by~\eqref{eq:GP Kalman filtering} for $ t\in (t_i, t_{i+1} )$, in which the subscript $t_+$ denotes the future trajectory from $t$ (see Fig.~\ref{fig:time_stamps}). This leads to an iterative framework which solves for the new approximation with up-to-date knowledge about the system.

We can apply this scheme to MDP/POMDP problems in both finite and receding horizon cases.\footnote{The receding-horizon version solves a new finite-horizon problem at each iteration of the Laplace approximation.} When facing a dynamic environment, \algo updates environmental information in the new Laplace approximation in Section~\ref{sec:Gaussian Density Approximate}. 

The details of the receding horizon approach are summarized in Algorithm~\ref{alg:algo} and can be derived similarly for the finite horizon case. First, at any time step $t_i$, \algo computes the Laplace approximation for the current horizon window $[t_i, t_i+t_h]$ with the latest information about the system and the environment, where $t_h \geq t_{i+1} - t_i$ is length of the preview horizon. Second, for $t \in (t_i, t_{i+1})$, \algo recursively updates the policy using the most current observation with a resolution of $\delta t$. These two steps repeat until the set criteria are met or the execution fails (for example, the robot is in collision).

%----------%
\begin{table*}[!t]
	\caption{Success rate across increasing $Q_x$ and $N_{obs}$ on the 2D holonomic robot.}
	\label{table:2dbenchmark}
	\begin{center}
		\begin{tabular}{|c|c||c|c||c|c||c|c||c|c||c|c|}
			\hline
			& \multirow{2}{*}{$\bf{Q_x}$} & \multicolumn{2}{c||}{\bf{10}} & \multicolumn{2}{c||}{\bf{20}} & \multicolumn{2}{c||}{\bf{30}} & \multicolumn{2}{c||}{\bf{40}} & \multicolumn{2}{c|}{\bf{50}}\\
			& & CL & OL & CL & OL & CL & OL & CL & OL & CL & OL \\
			\hline\hline
			\multirow{3}{*}{MDP}
			& 0.01 & 0.975 & 0.975 & 0.85 & 0.85 & 0.7 & 0.675 & 0.4 & 0.375 & 0.25 & 0.325 \\
			& 0.04 & 0.95 & 0.975 & 0.85 & 0.8 & 0.55 & 0.525 & 0.525 & 0.375 & 0.325 & 0.325 \\
			& 0.07 & 0.95 & 0.85 & 0.875 & 0.575 & 0.725 & 0.475 & 0.45 & 0.2 & 0.225 & 0.125 \\
			\hline
			\hline
			\multirow{3}{*}{POMDP}
			& 0.01 & 0.975 & 0.975 & 0.925 & 0.875 & 0.725 & 0.7 & 0.375 & 0.4 & 0.15 & 0.225 \\
			& 0.04 & 0.95 & 0.975 & 0.875 & 0.825 & 0.525 & 0.475 & 0.425 & 0.45 & 0.4 & 0.25 \\
			& 0.07 & 0.975 & 0.875 & 0.825 & 0.55 & 0.7 & 0.425 & 0.45 & 0.25 & 0.2 & 0.075 \\
			\hline 
		\end{tabular}
	\end{center}
\end{table*}

\begin{figure*}[!t]
	\centering
	\includegraphics[width=0.90\linewidth]{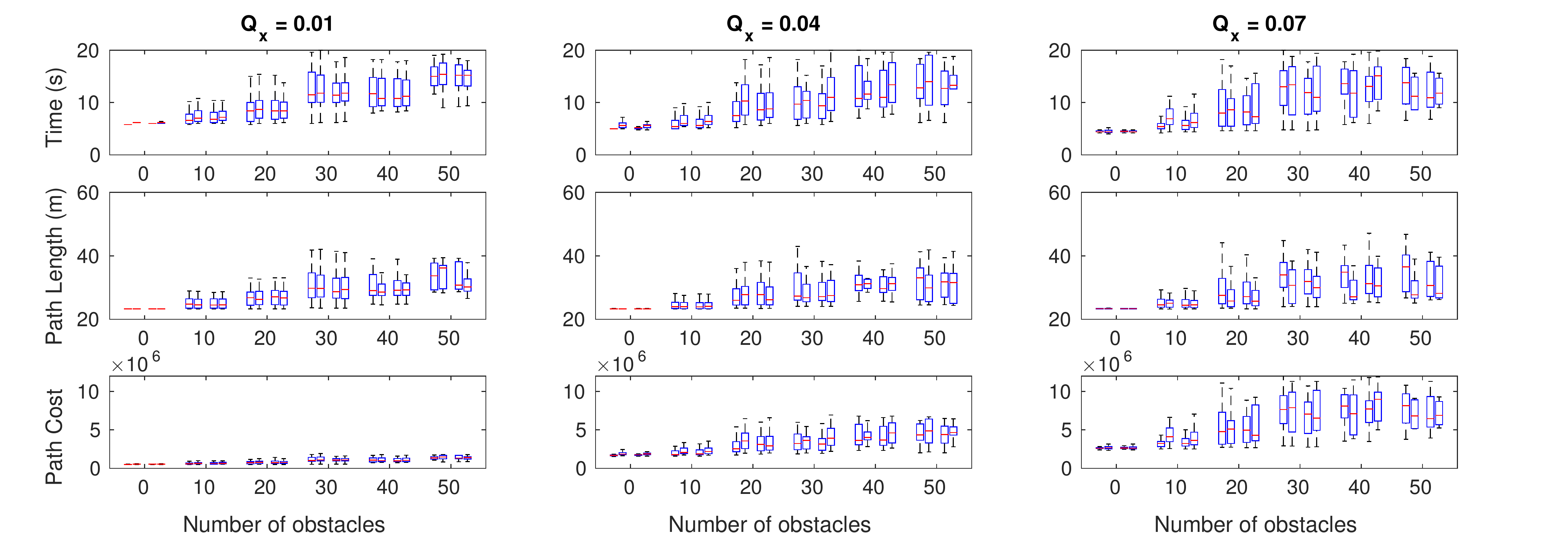}
	\caption{Results of successful runs with increasing $Q_x$ and $N_{obs}$ on the 2D holonomic robot.}
	\label{fig:2dbenchmark}
\end{figure*}
%----------%

%%%%%%%%%%%%%%%%%%%%%%%%%%%%%%%%%%%%%%%%%%%%%%%%%%%%%%%%%%%%%%%%%%%%%%%%%%%%%%%%%%%%%%%%%%%%%%%%%%%%%%%%%%%%%%%%

\vspace{-3mm}
\section{Implementation Details}\label{sec:imp}

We perform experiments with the receding horizon version of \algo in four different setups, including both MDP and POMDP scenarios: 
\mdpcl and \pomdpcl execute the receding horizon \algo in~Algorithm~\ref{alg:algo}; \mdpol and \pomdpol ignore the policy filtering step, but instead recursively apply the open-loop policy given as the mode found in the Laplace approximation. This open-loop baseline can be viewed as the direct generalization of~\cite{Dong-RSS-16} to include action trajectories. 

The Laplace approximation is implemented using GPMP2\footnote{Available at {https://github.com/gtrll/gpmp2}} and the GTSAM\footnote{Available at {https://bitbucket.org/gtborg/gtsam}} C++ library, which solves posterior maximization as a nonlinear least-squared optimization defined on a factor graph with the Levenberg-Marquardt algorithm.
Note that in implementation we consider $y_t = u_t$ (i.e. $\xi_t =  (x_t,u_t)$) and a constant time difference $\Delta t$ between any two support states or actions.

We evaluate our algorithms on three different systems: a 2D holonomic robot, a 7-DOF WAM, and a PR2 arm. The state dynamics, following~\eqref{eq:state SDE}, is defined as a double integrator with the state consisting of position and velocity and
\begin{equation*}
	\begin{small}
		A = \left[ \begin{matrix} \mathbf{0} & \mathbf{I} \\ \mathbf{0} & \mathbf{0} \end{matrix} \right],
		B = \left[ \begin{matrix} \mathbf{0} \\ \mathbf{I}  \end{matrix} \right],
		b = \left[ \begin{matrix} \mathbf{0} \\ \mathbf{0}  \end{matrix} \right],
		FF^T = \left[ \begin{matrix} \mathbf{0} & \mathbf{0} \\ \mathbf{0} & Q_x \mathbf{I} \end{matrix} \right]
	\end{small}
\end{equation*}
and following~\eqref{eq:control LTVSDE} we define the Gaussian process factor %for our case with $y=u$ i.e. considering only the first derivative of $u$ as a single integrator such that
by
\begin{equation*}
	\begin{small}
		H = \mathbf{0}, 
		D = \mathbf{0},
		\eta = 0,
		GG^T = Q_u \mathbf{I} 
	\end{small}
\end{equation*}
where $\mathbf{0}$ and $\mathbf{I}$ are $d \times d$ zero and identity matrices, where $d=2$ for the 2D holonomic robot and $d=7$ for the 7-DOF WAM arm and PR2 arm, and $Q_x$ and $Q_u$ are positive scalars. The observation process in the POMDP is modeled as a state observation with additive zero-mean Gaussian noise with covariance $Q_v = \sigma_m \mathbf{I}_{2d \times 2d}$. The state dynamics for both the arms are assumed to be feedback linearized. On a real system, the control would to be mapped back to real torques using inverse dynamics.

%%%%%%%%%%%%%%%%%%%%%%%%%%%%%%%%%%%%%%%%%%%%%%%%%%%%%%%%%%%%%%%%%%%%%%%%%%%%%%%%%%%%%%%%%%%%%%%%%%%%%%%%%%%%%%%%
\vspace{-1mm}
\section{Evaluation}\label{sec:eval}
We conduct benchmark experiments\footnote{A video of experiments is available at {https://youtu.be/8rQcg1O-6aU}} 
with our receding horizon algorithm on the 2D holonomic robot in a dynamic environment, and on the WAM arm and the PR2's right arm in a static environment (see Fig.~\ref{fig:misc}). In each case, we compare the closed-loop and open-loop algorithms for both MDP and POMDP settings across different $Q_x$ (and number of dynamic obstacles $N_{obs}$ in the 2D case) with respect to success rate, time to reach the goal, path length, and path cost.\footnote{Path cost is calculated as the negative log of the product of the factors.} Each setting is run for $K$ times with a unique random generator seed to account for stochasticity, which is kept the same across all four algorithms for a fair comparison. A trial is marked ``successful'' if the robot reaches the goal within a Euclidean distance $gdist$, and is marked ``failed'' if at any point the robot runs into collision or runs out of the maximum allotted time $t_{max}$. 

%------------------------------------------------------------------------------------------------------------------------------------------------%
\subsection{2D Robot Benchmark}
We simulate a 2D holonomic robot (radius $= 0.5m$) in a 2D environment ($30m \times 20m$) with moving obstacles (see Fig.~\ref{fig:misc} (a)). The robot's sensor returns a limited  view of a $5m \times 5 m$ square centered at the robot's current position.
The moving obstacles (squares of $1m \times 1m$) start at random locations and follow a 2D stochastic jump process, where a noisy acceleration $a_{obs}$ is uniformly sampled within $[-2.5, 2.5] m/s^2$ at every time step. Their velocities $v_{obs}$ are restricted within, $[-1.3, 1.3]m/s$. All obstacles are confined inside the boundary during simulation.

Table~\ref{table:2dbenchmark} summarizes the success rates for this benchmark,\footnote{Parameters for this benchmark are set as follows: $K=40$, $gdist = 0.2$, $t_{max} = 20$, $\Delta t = 0.2$, $t_h = 2$, $n_{ip} = 20$, $\sigma_m = 0.01$, $\sigma_g = 1$, $\sigma_{fix} = 10^{-4}$, $Q_u = 10$, $\sigma_{obs} = 0.02$, $\epsilon = 1$.} and Fig.~\ref{fig:2dbenchmark} shows the aggregate results of successful runs.
From Table~\ref{table:2dbenchmark}, we see that, for both MDP and POMDP cases, the closed-loop algorithms have higher success rates than the open-loop algorithms, especially in difficult problems with larger stochasticity in the system ($Q_x$) or increased complexity in the environment ($N_{obs}$). 
Similar increasing trends can also be observed in the difference of the success rates between the closed-loop and open-loop algorithms. 
The majority of failed open-loop cases arise from collision; only a few are due to hitting the maximum run time.
The performance in POMDP cases are slightly worse than that in the MDP cases on average. All three metrics (time, path length, and path cost) in Fig.~\ref{fig:2dbenchmark} increase in general with more noise and obstacles. It is important that these plots should be interpreted alongside the success rates, since the sample size of successful trails is comparatively sparse for the harder problems.

\begin{figure*}[!t]
	\centering
	\begin{subfigure}[b]{0.51\textwidth}
		\centering
		\includegraphics[width=0.84\linewidth]{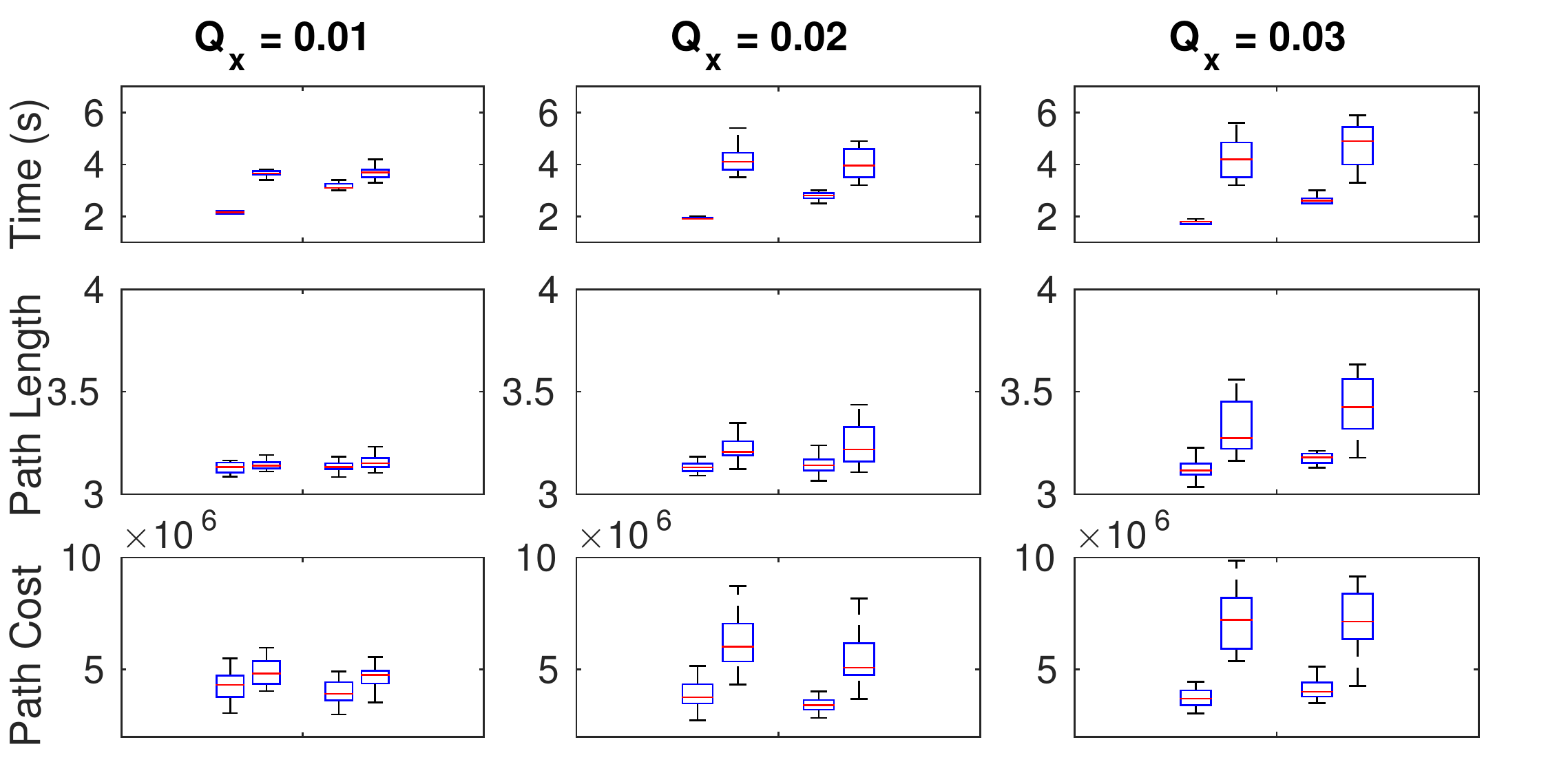}
		\vspace{-2mm}
		\caption{}
	\end{subfigure}
	\begin{subfigure}[b]{0.48\textwidth}
		\centering
		\includegraphics[width=0.84\linewidth]{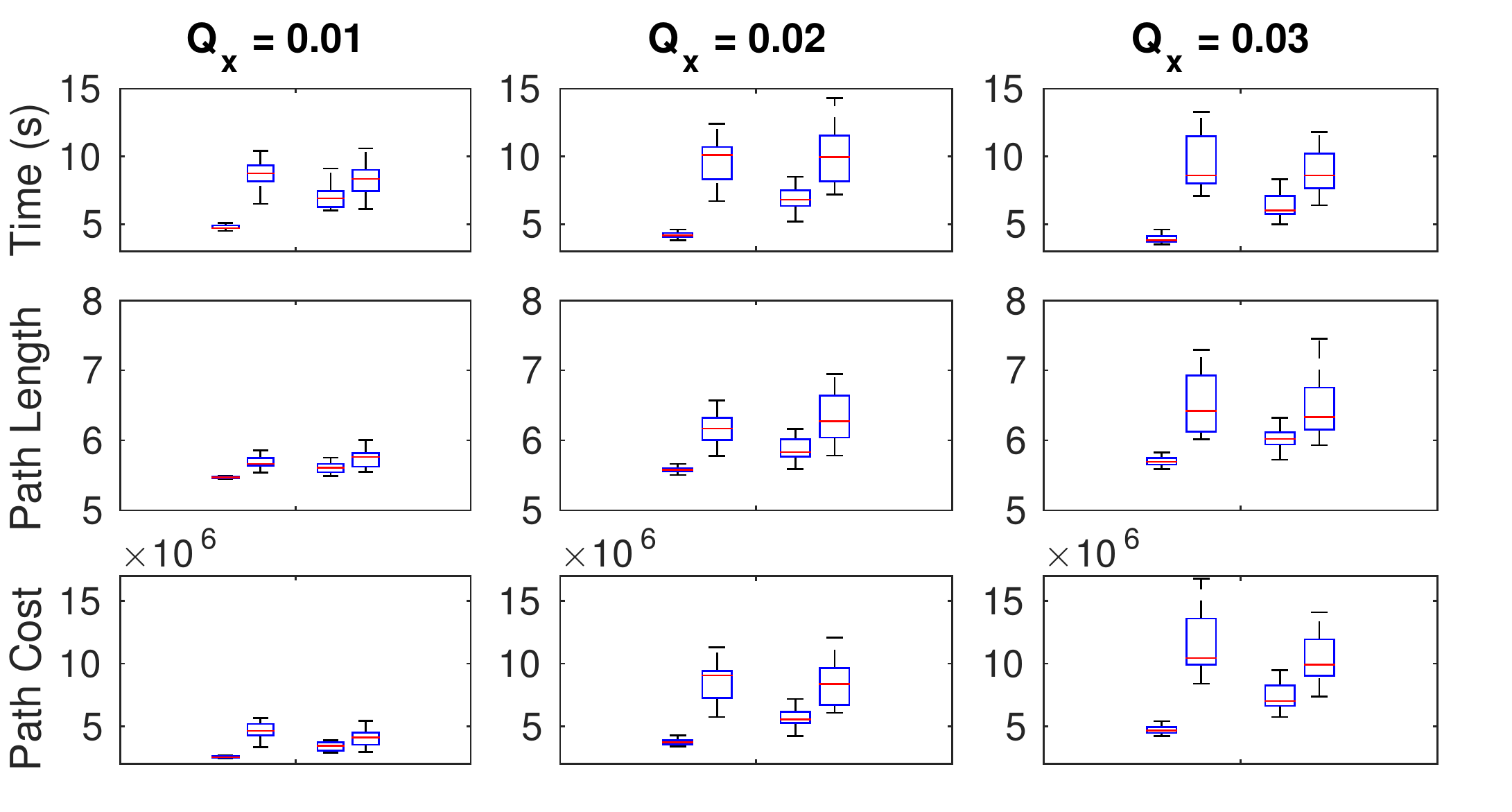}
		\vspace{-2mm}
		\caption{}
	\end{subfigure}
	\caption{Results of successful runs with increasing $Q_x$ on (a) the WAM and (b) the PR2 robot arms.}
	\label{fig:armplot}
	\vspace{-5mm}
\end{figure*}

\begin{table}[!t]
	\caption{Success rate across increasing $Q_x$ on the WAM and the PR2 robot arms.}
	\label{table:armsuccess}
	\begin{center}
		\begin{tabular}{|c|c||c|c||c|c|}
			\hline
			& \multirow{2}{*}{$\bf{Q_x}$} & \multicolumn{2}{c||}{\bf{WAM}} & \multicolumn{2}{c|}{\bf{PR2}} \\
			& & CL & OL & CL & OL \\
			\hline\hline
			\multirow{3}{*}{MDP}
			& 0.01 & 1   & 1       & 1    & 1   \\
			& 0.02 & 1   & 1       & 1    & 0.95 \\
			& 0.03 & 1   & 0.85  & 1    & 0.5 \\
			\hline
			\hline
			\multirow{3}{*}{POMDP}
			& 0.01 & 1    & 1      & 1 & 1   \\
			& 0.02 & 1    & 0.9   & 1 & 0.8 \\
			& 0.03 & 0.9 & 0.75 & 1 & 0.8 \\
			\hline 
		\end{tabular}
	\end{center}
	\vspace{-5mm}
\end{table}

%------------------------------------------------------------------------------------------------------------------------------------------------%
\subsection{WAM and PR2 Benchmark}

We demonstrate the scalability of \algo to higher dimensional systems by performing a benchmark on the WAM and the PR2 robot arms.
Here the WAM and the PR2 robot arms are set up in \emph{lab} and \emph{industrial} environments~\cite{Mukadam-ICRA-16,Dong-RSS-16,schulman2014motion} respectively, in OpenRAVE. Here the task is to drive the robot arm from a given start to a goal configuration (see Fig.~\ref{fig:misc} (b) and (c)). The environments are static and fully observable at all times. We compare 
the algorithms with respect to increasing $Q_x$.
Table~\ref{table:armsuccess} summarizes the success rates for this benchmark,\footnote{Parameters for this benchmark are set as follows: $K=20$, $gdist = 0.065$, $t_{max} = 15$, $\Delta t = 0.1$, $t_h = 2$, $n_{ip} = 10$, $\sigma_m = 0.005$, $\sigma_g = 0.03$, $\sigma_{fix} = 10^{-4}$, $Q_u = 10$, $\epsilon = 0.1$, $\sigma_{obs} = 0.008$ (WAM), $\sigma_{obs} = 0.005$ (PR2).} and Fig.~\ref{fig:armplot} shows the aggregate results of successful runs. 
Similar to the 2D robot benchmark, the results show that the closed-loop algorithms have higher success rate than the open-loop ones, and all three metrics increase  with noise. In particular, \pomdpcl performs even better than MDP-OL.

%%%%%%%%%%%%%%%%%%%%%%%%%%%%%%%%%%%%%%%%%%%%%%%%%%%%%%%%%%%%%%%%%%%%%%%%%%%%%%%%%%%%%%%%%%%%%%%%%%%%%%%%%%%%%%%%
\section{Conclusion}\label{sec:conc}
We consider the problem of motion planning and control as probabilistic inference, and we propose an algorithm \algo for solving this problem that can exploit intrinsic sparsity in continuous-time stochastic systems. In particular, \algo can address performance indices given by arbitrary, higher-order nonlinear factors and a general exponential-integral-quadratic factor. Despite \algo solving a continuous-time problem, its complexity scales only linearly in the number of nonlinear factors, thus making online simultaneous planning and control possible in receding/finite horizon MDP/POMDP problems.
\vspace{-3mm}

%%%%%%%%%%%%%%%%%%%%%%%%%%%%%%%%%%%%%%%%%%%%%%%%%%%%%%%%%%%%%%%%%%%%%%%%%%%%%%%%%%%%%%%%%%%%%%%%%%%%%%%%%%%%%%%%
\section*{Acknowledgments}
The authors would like to thank Jing Dong for help with the GTSAM interface. This material is based upon work supported by NSF CRII Award No. 1464219 and NSF NRI Award No. 1637758.
\vspace{-3mm}

%%%%%%%%%%%%%%%%%%%%%%%%%%%%%%%%%%%%%%%%%%%%%%%%%%%%%%%%%%%%%%%%%%%%%%%%%%%%%%%%%%%%%%%%%%%%%%%%%%%%%%%%%%%%%%%%

\bibliographystyle{IEEEtran}
\bibliography{IEEEabrv,ref}

%%%%%%%%%%%%%%%%%%%%%%%%%%%%%%%%%%%%%%%%%%%%%%%%%%%%%%%%%%%%%%%%%%%%%%%%%%%%%%%%%%%%%%%%%%%%%%%%%%%%%%%%%%%%%%%%

\clearpage

\section*{Appendix}

%------------------------------------------------------------------------------------------------------------------------------------------------%
\subsection{Laplace Approximation with Factor Graphs}

\algo updates the Laplace approximation whenever $t=t_i$ by efficiently solving a nonlinear least-squares problem defined on a bipartite \emph{factor graph} $\mathcal{G} = \{\bm{\xi}_\SS, \bm{f}_\SS, \mathcal{E} \}$,
\begin{equation}
q(\bm{\xi}_{\SS} | \bm{e}_\SS) \propto \prod \limits_{S\in \SS} f_{S} (\xi_S). \label{eq:factor_graph}
\end{equation}
where recall that $\bm{\xi}_\SS$ is the set of support augmented states, and $\bm{f}_\SS= \{f_S \}_\SS$ denotes the set of factors, and $\mathcal{E}$ are edges connected to between  $\bm{\xi}_\SS$ and $\bm{f}_\SS$. 

An example factor graph is shown in Fig.~\ref{fig:factor_graph} for a trajectory starting from $t_i$ with a length equal to  $t_h$. The sparse set of support augmented states $\bm{\xi}_\SS$ are uniformly $\Delta t$ apart and are connected to their neighbours via the Gaussian process factors, forming a Gauss-Markov chain. 
Note that in our implementation $\xi_t = (x_t,u_t)$.

\subsubsection{Details of Factor Implementation}
\paragraph{Prior Factor}
For each Laplace approximation, a prior factor is placed on the first hidden state $\xi_t$, reflecting its current belief given past history $\bm{h}_t$. In the MDP setting, the covariance for state $x_t$ is set as a diagonal matrix $\sigma_{fix}^2 \mathbf{I}_{2d \times 2d}$, in which $\sigma_{fix}$ is a small number to indicate high confidence; for control $u_t$, we use the original Gaussian process factor given by~\eqref{eq:gp factor}. Together they define $Q_{prior}$. In the POMDP setting, the belief of the hidden augmented state is obtained via Kalman filtering, and we heuristically set the covariance for the state, $x$ to $\sigma_{fix}^2 \mathbf{I}_{2d \times 2d}$ as mentioned previously.

\begin{figure}[!t]
	\begin{centering}
		{\includegraphics[width=0.82\columnwidth]{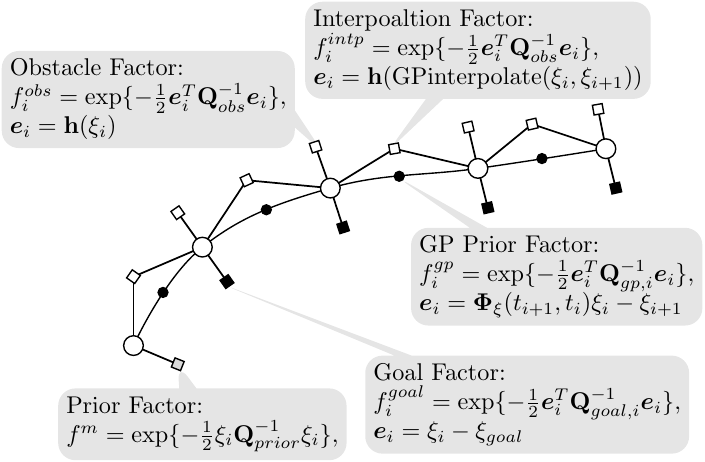}}
		\par\end{centering}
	\protect\caption{A factor graph of an example Laplace approximation problem showing states (white circle) ($\xi_i$ is used as a shorthand for $\xi_{t_i}$) and different kinds of factors: GP Prior (black circle), obstacle and interpolation (white square), measurement (gray square) and goal (black square).
		\label{fig:factor_graph}}
\end{figure}

\paragraph{Gaussian Process Factors}
Analogous to defining $\GP_u$ for~\eqref{eq:control LTVSDE}, we can define $\GP_{\xi}$, which in turns define. $q (\bm{\xi}_\SS )$ in~\eqref{eq:approximate conditional marginal}.
In Fig.~\ref{fig:factor_graph}, this corresponds to Gaussian process factors with
\begin{equation*}
	\mathbf{Q}_{gp,i} = \int_{t_{i}}^{t_{i+1}} \Phi_\xi(t_{i+1},s) \left[ \begin{matrix} F \\ G  \end{matrix} \right] \left[ \begin{matrix} F \\ G  \end{matrix} \right]^T \Phi_\xi(t_{i+1},s)^T ds,
\end{equation*}
where $\Phi_\xi$ is the state transition matrix associated with $\left[ \begin{smallmatrix} A & B \\ \mathbf{0} & H  \end{smallmatrix} \right]$ that takes the system from $t_{i}$ to $t_{i+1}$.

\paragraph{Obstacle and Interpolation Factors}
For obstacle avoidance, we use a hinge loss function $\mathbf{h}$ with safety distance $\epsilon$ to compute a signed distance field as in GPMP2~\cite{Dong-RSS-16}. In effect, it defines the obstacle factors and interpolation factors in Fig.~\ref{fig:factor_graph}, which both use $Q_{obs} = \sigma_{obs}^2 \mathbf{I}$. Though abstracted as a single factor in Fig.~\ref{fig:factor_graph}, between any two support points $t_{i}$ and $t_{i+1}$, multiple ($n_{ip}$) interpolated factors can be constructed with indexes evenly spaced in time ($\delta t$ apart) to ensure path safety. See~\cite{Dong-RSS-16} for details.

\paragraph{Goal Factor}
To drive the system to a desired goal configuration $\xi_{goal}$ (for example, a particular position in configuration space with zero velocity and action), we add a goal factor to every support point except the current state. 
This encourages the optimizer to drive all states in the current horizon window closer to the goal.
Because the covariance of this factor acts as a weighting term, we define it as	
$	\mathbf{Q}_{goal,i} = \sigma_{g}^2 \frac{||\xi_{t_i} - \xi_{goal}||^2}{||\xi_{start} - \xi_{goal}||^2} \mathbf{I}$
such that it monotonically decreases with the Euclidean distance to the goal.

\subsubsection{Update of Laplace Approximation}
The same Laplace approximation is used to recursively update the policy for $t \in [t_i, t_{i+1})$ with a resolution of $\delta t$, and, when $ t = t_{i+1}$, the graph is updated to construct a new nonlinear least-square optimization problem. 
This is done by shifting the horizon window ahead by $\Delta t$ and update the factors to include any environmental changes  during $[t_i, t_{i+1})$. In the updated graph, the prior factor on the first state 
is given by an additional Kalman filter based on~\eqref{eq:state SDE} and~\eqref{eq:control LTVSDE} with $(x,y)$ as hidden states and $(z,u)$ as observations. For POMDP problems, we treat the estimation of current state as perfect knowledge without uncertainty. This extra heuristic step is a compromise which makes the assumption accurate at the mean of the current belief.

%------------------------------------------------------------------------------------------------------------------------------------------------%
\subsection{Proof of~\eqref{eq:determinisitc policy}}

\noindent Here we prove that the solution~\eqref{eq:determinisitc policy} to the approximate optimization problem~\eqref{eq:LEQG approx II}
\begin{align*}
\max_{\bm{\pi} } \int  \hat{q}(\bm{z}, \bm{\xi} | \bm{e}_\SS)  \bm{\pi}(\bm{u} | \bm{h}) \der \bm{\xi} \der \bm{z} 
\end{align*}
can be written as posterior inference: 
$\forall t$,
\begin{align*}
\pi_t^*( u_t | \bm{h}_t ) &= \delta(u_t - u_t^*(\bm{h}_t)) \nonumber  \\
u_t^*(\bm{h}_t) &= \argmax_{u_t}  \int  \hat{q}(\bm{z},\bm{\xi} | \bm{e}_\SS )  \bm{\pi}^*( \bar{\bm{u}}_t | \bm{h}) \der \bm{x} \der \bm{y}  \der \bm{z} \der \bar{\bm{u}}_t  \nonumber \\
&=\argmax_{u_t} \hat{q}(u_t| \bm{h}_t, \bm{e}_\SS) 
\end{align*}
where $ \hat{q}(\bm{z}, \bm{\xi} | \bm{e}_\SS) \propto  q(\bm{\xi})  p(\bm{z}|\bm{x}) \hat{p}(\bm{e}_\SS | \bm{x}_\SS, \bm{u}_\SS )$ and $ \hat{p}(\bm{e}_\SS | \bm{x}_\SS, \bm{u}_\SS )  = \prod_{S\in\SS} \hat{p}(e_S | x_S, u_S )$ is found by the exponential-quadratic approximate factor given by the Laplace approximation.

\begin{proof}
We assume the length of the trajectory is $T$.  In the following, we first show that the optimal policy is deterministic and then show that it corresponds to the mode of the posterior distribution $\hat{q}(u_t| \bm{h}_t, \bm{e}_\SS)$. 

\paragraph{The Optimal Policy is Deterministic}

For any $t$, we can write the objective function~\eqref{eq:LEQG approx II} as 
\begin{align*}
&\int  \hat{q}(\bm{z}, \bm{\xi} | \bm{e}_\SS)  \bm{\pi}(\bm{u} | \bm{h}) \der \bm{\xi} \der \bm{z} \\
&\propto 
\int   q(\bm{\xi})  p(\bm{z}|\bm{x}) \hat{p}(\bm{e}_\SS | \bm{x}_\SS, \bm{u}_\SS )  \bm{\pi}(\bm{u} | \bm{h}) \der \bm{\xi} \der \bm{z} \\
&= \int  \pi_{t}(u_t | \bm{h}_t ) f_{\bm{h}_t}(u_t) \der u_t   \der \bm{h}_t
\end{align*}
in which 
\begin{align*}
f_{\bm{h}_t}(u_t)  = \int  q(\bm{\xi})  p(\bm{z}|\bm{x}) \hat{p}(\bm{e}_\SS | \bm{x}_\SS, \bm{u}_\SS ) \bm{\pi}(  \bar{\bm{u}}_t | \bar{\bm{h}}_t) \der \bm{\theta} \der \bm{z}_{t_+} \der {\bm{u}}_{t_+}
\end{align*}
and 
\[ \bm{\pi}(  \bar{\bm{u}}_t | \bar{\bm{h}}_t) := \frac{\bm{\pi}(\bm{u} | \bm{h} )}{\pi_{t}(u_t | \bm{h}_t )}\]
Therefore, equivalently,~\eqref{eq:LEQG approx II} can be formulated  explicitly as a variational problem of density function $\pi_t$:
\begin{align}
& \qquad 
\max_{\pi_t} \int  \pi_{t}(u_t | \bm{h}_t ) f_{\bm{h}_t}(u_t) \der u_t   \label{eq:var problem I} \\
& s.t. \nonumber \\
& \qquad   \int  \pi( u_t | \bm{h}_t)  \der u_t = 1  \nonumber \\
& \qquad    \pi( u_t | \bm{h}_t )   \geq  0, \quad \forall u_t  \nonumber
\end{align}
To deal with the inequality, let $g_t^2(u_t) = \pi_t(u_t | \bm{h}_t)$, and we can further write~\eqref{eq:var problem I} as 
\begin{align}
& \qquad 
\max_{g_t}\int  g_t^2(u_t) f_{\bm{h}_t}(u_t) \der u_t  \label{eq:var problem II}  \\
& s.t. \nonumber \\
& \qquad   \int  g_t^2(u_t) \der u_t = 1 \nonumber
\end{align}
Let $\lambda \in \R$ be a Lagrangian multiplier. We can turn the~\eqref{eq:var problem II} into an unconstrained optimization and use calculus of variations to derive the solution:
\begin{align*}
\min_\lambda&  \max_{g_t}  \LL(g_t, \lambda) \\
	&= \min_\lambda  \max_{g_t} \int  g_t^2(u_t) f_{\bm{h}_t}(u_t) \der u_t
	+ \lambda (\int  g_t^2(u_t)  \der u_t - 1)  
\end{align*}
Suppose $g_t^*(\lambda)$ is the optimum. Let $g_t = g_t^* + \epsilon \eta$, where $\eta$ is an arbitrary continuous function and $\epsilon \rightarrow 0 $.  
Then the optimality condition is given by 
\begin{align*}
\frac{ \partial  \LL(g_t, \lambda)} {\partial \epsilon} = 
 \int  2 g_t(u_t) \eta(u_t) (  \lambda + f_{\bm{h}_t}(u_t)) \der u_t =0.
\end{align*}
Since $\eta$ is arbitrary, it implies $\forall u_t$,
\[
g_t(u_t)(  \lambda + f_{\bm{h}_t}(u_t)) = 0
\]
Given that $\lambda$ is a scalar and $g(u_n)$ is non-zero, 
we can conclude that $ \pi_t^*(u_t|\bm{h}_t) = \delta(u_t - u_t^*(\bm{h}_t))$ satisfying
\begin{align*}
u_t^*( \bm{h}_t) = \arg \max_{u_t( \bm{h}_t)} f_{\bm{h}_t}(u_{t})
\end{align*}

\paragraph{The Optimal Policy is the Mode of Posterior}

From the previous proof, we know that the policy corresponds to the mode of $f_{\bm{h}_t}(u_{t})$ for any $t$. Therefore, to complete the proof, we only need to show that $f^*_{\bm{h}_t}(u_{t}) \propto \hat{q}(u_t | \bm{e}_\SS , \bm{h}_t )$, where $f^*_{\bm{h}_t}(u_{t})$ is $f_{\bm{h}_t}(u_{t}) $ when the policies are optimal.

First, let $\hat{f}_{\bm{h}_t}(u_t)$ denote $ f_{\bm{h}_t}(u_t)$ when all policies are deterministic, and define, for all $t$,
\begin{align}
\hat{f}_{\bm{h}_t}(u_t) &:= \bm{\pi}_{t_-} (\bm{u}_{t_-}|\bm{h}_{t_-})  \hat{q}( \bm{z}_t, \bm{z}_{t_-}, \bm{e}_\SS , \bm{u}_{t_-}, u_t  ) \label{eq:optimal f}
\end{align}
Next, we introduce a lemma:
\begin{lemma} \label{lm:dualtiy}
Let $\bm{z} = (x,y) \in \R^n$. If $f(x,y) \propto 
\NN \left(
\bm{z}| 
\bm{m}, \bm{S}
\right)$, then, for all $y$, 
\begin{align}
\max_{x(y)} f(x,y)  = C \int f(x,y) \der x \label{eq:lemma}
\end{align}
for some constant $C$ independent of $y$, in which $\bm{m}$ and $\bm{S}$ are the mean and covariance of a Gaussian. 
\end{lemma}

Now we can show $f^*_{\bm{h}_t}(u_{t}) \propto \hat{q}(u_t | \bm{e}_\SS, \bm{h}_t ) $ by induction in backward order. To start with, for the last policy at $ \tau = T - \delta t$, we can write~\eqref{eq:optimal f} as
\begin{align*}
f_{\bm{h}_{\tau} }( u_{\tau}) & = \hat{f}_{\bm{h}_{\tau} }( u_{\tau})= 
\bm{\pi}_{\tau_-} (\bm{u}_{\tau_-}|\bm{h}_{\tau_-})  \hat{q}( \bm{z}_\tau, \bm{z}_{\tau_-},  \bm{e}_\SS, \bm{u}_{\tau_-}, u_\tau  )
\\
&=\bm{\pi}_{\tau_-} (\bm{u}_{\tau_-}|\bm{h}_{\tau_-}) \hat{q}( \bm{h}_\tau,  \bm{e}_\SS, u_\tau  ) \\
&\propto \hat{q}( u_\tau | \bm{h}_\tau,  \bm{e}_\SS  )
\end{align*}
in which we purposefully omit the dependency of $\bm{u}_{\tau_-}$ on $\bm{h}_{\tau_-}$, because the exact value of  $\bm{u}_{\tau_-}$ is observed when performing the optimization.
Then we have  $u^*_\tau(\bm{h}_\tau) = \argmax_{u_\tau} \hat{f}_{\bm{h}_\tau}(u_\tau) = \argmax_{u_\tau}  \hat{q}( u_\tau | \bm{h}_\tau,  \bm{e}_\SS  )$.

Now we propagate the objective function one step backward from $\tau $ to $\tau - \delta t$. Given $\bm{h}_{\tau - \delta t}$, the maximization at $\tau - \delta t$ is given as
\begin{align*}
&\max_{\pi_{\tau - \delta t}}\max_{\pi_\tau} \int  \pi_{\tau}(u_\tau | \bm{h}_\tau ) \hat{f}_{\bm{h}_\tau}(u_\tau) \der u_\tau  \der z_\tau  \der u_{\tau - \delta t} \\
&=\max_{\pi_{\tau - \delta t}}\int \max_{u_{\tau}(\bm{h}_{\tau})} \hat{f}_{\bm{h}_\tau}(u_\tau)  \der z_\tau \der u_{\tau - \delta t}  \\
&=\max_{\pi_{\tau - \delta t}} \int \bm{\pi}_{\tau_-} (\bm{u}_{\tau_-}|\bm{h}_{\tau_-})  \max_{u_{\tau}(\bm{h}_{\tau})} \hat{q}( \bm{h}_{\tau}, \bm{e}_\SS, u_{\tau} )   \der z_\tau  \der u_{\tau - \delta t} \\
&\propto \max_{\pi_{\tau - \delta t}} \int \bm{\pi}_{\tau_-} (\bm{u}_{\tau_-}|\bm{h}_{\tau_-})  \hat{q}( \bm{h}_{\tau}, \bm{e}_\SS, u_{\tau} )   \der z_\tau \der u_{\tau}  \der u_{\tau - \delta t}\\
&=\max_{\pi_{\tau - \delta t}}  \int \bm{\pi}_{\tau_-} (\bm{u}_{\tau_-}|\bm{h}_{\tau_-})  \hat{q}( \bm{h}_{\tau-\delta t}, \bm{e}_\SS, u_{\tau - \delta t} )   \der u_{\tau - \delta t} \\
&= \max_{\pi_{\tau - \delta t}} \int \pi_{\tau - \delta t}( u_{\tau - \delta t} | \bm{h}_{\tau - \delta t}) 
 \hat{f}_{\bm{h}_{\tau- \delta t}}(u_{\tau - \delta t})  \der u_{\tau - \delta t}
\end{align*}
The second equality is due to the policy is deterministic; the third proportionality is given by~\eqref{eq:lemma}; the last equality is given by the definition~\eqref{eq:optimal f}. 
Therefore, the backward iteration maintains the policy optimization problem 
$\max_{\pi_t} \int  \pi_{t}(u_t | \bm{h}_t ) \hat{f}_{\bm{h}_t}(u_t) \der u_t$
in the same algebraic form as the last step for all $t$. 
Since $\hat{f}_{\bm{h}_t}(u_t) \propto \hat{q}( u_t | \bm{h}_t,  \bm{e}_\SS  )$, this completes the proof. 

\end{proof}

\end{document}